\def\year{2017}\relax
\documentclass[letterpaper]{article}
\usepackage{aaai17}
\usepackage{times}
\usepackage{helvet}
\usepackage{courier}
\usepackage{graphicx}
\usepackage{wrapfig}
\usepackage{url}
\usepackage{amsmath}
\usepackage[font={small}]{caption}
\usepackage{color}
\usepackage{amsmath,amssymb}
\usepackage[font={small}]{caption}
\usepackage{enumitem}
\usepackage{bm}
\usepackage{color,xcolor,colortbl}

\DeclareMathOperator*{\argmax}{arg\,max\,}
\newcommand\numberthis{\addtocounter{equation}{1}\tag{\theequation}}

\title{Toward Extractive Summarization of Online Forum Discussions \\via Hierarchical Attention Networks}

\author{Sansiri Tarnpradab \quad Fei Liu \quad Kien A. Hua \\
Department of Computer Science \\
University of Central Florida, Orlando, FL 32816 \\
  {\tt sansiri@knights.ucf.edu} \quad {\tt \{feiliu, kienhua\}@cs.ucf.edu} \\}

\date{}

\begin{document}
\maketitle
\begin{abstract}
Forum threads are lengthy and rich in content.
Concise thread summaries will benefit both newcomers seeking information and those who participate in the discussion.
Few studies, however, have examined the task of forum thread summarization.
In this work we make the first attempt to adapt the hierarchical attention networks for thread summarization.
The model draws on the recent development of neural attention mechanisms to build sentence and thread representations and use them for summarization.
Our results indicate that the proposed approach can outperform a range of competitive baselines.
Further, a redundancy removal step is crucial for achieving outstanding results.

\end{abstract}

\section{Introduction}
\label{sec:intro}


Online forums play an important role in shaping public opinions on a number of issues, ranging from popular tourist destinations to major political events.
As a form of new media, the influence of forums is on the rise and rivals that of traditional media outlets~\cite{Stephen:2012}.
A forum thread is typically initiated by a user posting a question or comment through the website. 
Others reply with clarification questions, further details, solutions, and positive/negative feedback~\cite{Bhatia:2014}.
This corresponds to a community-based knowledge creation process where knowledge of enduring value is preserved~\cite{Anderson:2012}.
It is not uncommon that forum threads are lengthy and comprehensive, containing hundreds of pages of discussion.
In this work we seek to generate concise forum thread summaries that will benefit both the newcomers seeking information and those who participate in the discussion.

Few studies have examined the task of forum thread summarization.
Traditional approaches are largely based on multi-document summarization frameworks.
Ding and Jiang~\shortcite{Ding:2015} presented a preliminary study on extracting opinionated summaries for online forum threads. 
They analyzed the discriminative power of a range of sentence-level features, including relevance, text quality and subjectivity. 
Bhatia et al.~\shortcite{Bhatia:2014} studied the effect of dialog act labels on predicting summary posts.
They define a thread summary as a collection of relevant posts from a discussion.
Ren et al.~\shortcite{Ren:2011} approached the problem using hierarchical Bayesian models and performed random walks on the graph to select summary sentences.
The aforementioned studies used datasets ranging from 10 to 400 threads. 
Due to the lack of annotated datasets, supervised summarization approaches have largely been absent from this space.

In this work we introduce a novel supervised thread summarization approach that is adapted from the hierarchical attention networks (HAN) proposed in~\cite{Yang:2016}.
The model draws on the recent development of neural attention mechanisms.
It learns effective sentence representation by attending to important words, and similarly learns thread representation by attending to important sentences in the thread.
Hierarchical network structures have seen success in both document modeling~\cite{Li:2015} and machine comprehension~\cite{Yin:2016}. 
To the best of our knowledge, this work is the first attempt to adapt it to forum thread summarization. 
We further created a dataset by manually annotating 600 threads with human summaries. 
The annotated data allow the development of a supervised system trained in an end-to-end fashion. 
We compare the proposed approach against state-of-the-art summarization baselines.
Our results indicate that the HAN models are effective in predicting summary sentences.
Further, a redundancy removal step is crucial for achieving outstanding results.


\section{Our Approach}
\label{sec:framework}

We formulate thread summarization as a task that extracts relevant sentences from a discussion.
A sentence is used as the extraction unit due to its succinctness. 
The task naturally lends itself to a supervised learning framework. 
Let $\bm{s}$=$[s_1,\cdots,s_N]$ be the sentences in a thread
and $\bm{t}$=$[t_1,\cdots,t_N]$ be the binary labels, where 1 indicates the sentence is in the summary and 0 otherwise.
The task of forum thread summarization is to find the most probable tag sequence given the thread sentences: 
\begin{align*}
\argmax_{\bm{t} \in \mathcal{T}} p(\bm{t}|\bm{s}) \numberthis
\end{align*}
\noindent where $\mathcal{T}$ is the set of all possible tag sequences.
In this work we make independent tagging decisions, where $\textstyle p(\bm{t}|\bm{s}) = \prod_{i=1}^N p(t_i|\bm{s})$.
We begin by describing the hierarchical attention networks (HAN; Yang et al., 2016)\nocite{Yang:2016}
that are used to construct sentence and thread representations,
followed by our adaptation of the HAN models to thread summarization.
Below we use bold letters to represent vectors and matrices (e.g., $\bm{h}_t, \bm{W}$).
Words and sentences are denoted by their indices.


\vspace{0.05in}
\noindent\textbf{Sentence Encoder.} 
It reads an input sentence and outputs a sentence vector. 
Inspired by recent results in~\cite{Bahdanau:2015,Chen:2016}, we use a bi-directional recurrent neural network as the sentence encoder.
The model additionally employs an attention mechanism that learns to attend to important words in the sentence while generating the sentence vector.

Let $\bm{s}_i$=$[x_1,\cdots,x_T]$ be the $i$-th sentence and the words are indexed by $t$.
Each word is replaced by a pretrained word embedding before it is fed to the neural network. 
We use the 300-dimension word2vec embeddings~\cite{mikolov-13} pretrained on Google News dataset with about 100 billion words.
While both gated recurrent units (GRU, Chung et al., 2014)\nocite{Chung:2014} and long short-term memory (LSTM, ochreiter and Schmidhuber 1997)\nocite{hochreiter-97} are variants of recurrent neural networks, we opt for LSTM in this study due to its proven effectiveness in previous studies.

LSTM embeds each word into a hidden representation $\bm{h}_t$=LSTM$(\bm{h}_{t-1},\bm{x}_t)$.
It employs three gating functions (input gate $\bm{i}_t$ (Eq.(2)), forget gate $\bm{f}_t$ (Eq.(3)), and output gate $\bm{o}_t$ (Eq.(4))) to control how much information comes from the previous time step, and how much will flow to the next.  
The gating mechanism is expected to keep information flow for a long period of time.
In particular, Eq.(6) calculates the cell state $\bm{C}_{t}$ by selectively inheriting information from $\tilde{\bm{C}_t}$ (via the input gate) and from $\bm{C}_{t-1}$ (via the forget gate). 
Eq.(7) generates the hidden state by applying the output gate to $\tanh(\bm{C}_{t})$. The equations are described below.
\begin{align*}
\bm{i}_t &= \sigma(\bm{W}^i\bm{x}_t + \bm{U}^i\bm{h}_{t-1} + \bm{b}^i) \numberthis\\
\bm{f}_t &= \sigma(\bm{W}^f\bm{x}_t + \bm{U}^f\bm{h}_{t-1} + \bm{b}^f) \numberthis\\
\bm{o}_t &= \sigma(\bm{W}^o\bm{x}_t + \bm{U}^o\bm{h}_{t-1} + \bm{b}^o) \numberthis\\
\tilde{\bm{C}_t} &= \tanh(\bm{W}^c\bm{x}_t + \bm{U}^c\bm{h}_{t-1} + \bm{b}^c) \numberthis\\
\bm{C}_t &= \bm{i}_t \odot \tilde{\bm{C}_t} + \bm{f}_t \odot \bm{C}_{t-1} \numberthis\\
\bm{h}_t &= \bm{o}_t \odot \tanh(\bm{C}_t) \numberthis
\end{align*}



\noindent where $\odot$ is the element-wise product of two vectors.
We additionally employ a bi-directional LSTM model that includes a forward-pass (Eq.(8)) and a backward pass (Eq.(9)). 
$\overrightarrow{\bm{h}_t}$ is expected to carry over semantic information from beginning of the sentence to the current time step; 
whereas $\overleftarrow{\bm{h}_t}$ encodes information from the current time step to the end of sentence.
Concatenating the two vectors $\bm{h}_t$=$[\overrightarrow{\bm{h}_t}, \overleftarrow{\bm{h}_t}]$ produces a word representation that encodes the sentence-level context.
\begin{align*}
\overrightarrow{\bm{h}_t}=\text{LSTM}_1(\overrightarrow{\bm{h}_{t-1}}, \bm{x}_t) \numberthis\\
\overleftarrow{\bm{h}_t}=\text{LSTM}_2(\overleftarrow{\bm{h}_{t-1}}, \bm{x}_t) \numberthis
\end{align*}


Next we describe the attention mechanism. 
Of key importance is the introduction of a vector $\bm{u}_w$ for all words, which is \emph{trainable} and expected to capture ``global'' word saliency. We first project $\bm{h}_t$ to a transformed space and generates $\bm{u}_t$ (Eq.(10)).
The inner product $\bm{u}_t^T \bm{u}_w$ is expected to signal the importance of the $t$-th word.
It is converted to a normalized weight $\alpha_t$ through a softmax function (Eq.(11)).
\begin{align*}
\bm{u}_{t} &= \tanh(\bm{W}^a \bm{h}_{t}+\bm{b}^a) \numberthis\\
\alpha_{t} &= \frac{\exp(\bm{u}_t^T \bm{u}_w)}{\sum_{t} \exp(\bm{u}_t^T \bm{u}_w)} \numberthis
\end{align*}
The sentence vector $\bm{s}_i$ is generated as a weighted sum of word representations, where $\alpha_t$ is a scalar value indicating the word importance (Eq.(12)).
\begin{align*}
\bm{s}_i = \sum_t \alpha_t \bm{h}_t \numberthis
\end{align*}

\noindent\textbf{Thread Encoder.}
It takes as input a sequence of sentence vectors $\bm{s}$=$[\bm{s}_1,\cdots,\bm{s}_N]$ encoded using the sentence encoder described above and outputs a thread vector.
Assume the sentences are indexed by $i$.
The thread encoder employs the same network architecture as the sentence encoder. We summarize the equations below.
Note that the attention mechanism additionally introduces a vector $\bm{u}_s$ for all sentences, which is \emph{trainable} and encodes salient sentence-level content.
The thread vector $\bm{s}$ is a weighted sum of sentence vectors, where $\alpha_i$ is a scalar value indicating the importance of the $i$-th sentence.
\begin{align*}
\overrightarrow{\bm{h}_i} &= \text{LSTM}_3(\overrightarrow{\bm{h}_{i-1}},\bm{s}_{i}) \numberthis\\
\overleftarrow{\bm{h}_i} &= \text{LSTM}_4(\overleftarrow{\bm{h}_{i-1}}, \bm{s}_{i}) \numberthis\\
\bm{h}_{i} &= [\overrightarrow{\bm{h}_{i}} , \overleftarrow{\bm{h}_{i}}] \numberthis\\
\bm{u}_{i} &= \tanh(\bm{W}^b \bm{h}_{i} + \bm{b}^b) \numberthis\\
\alpha_{i} &= \frac{\exp(\bm{u}_{i}^T \bm{u}_{s})}{\sum_{i} \exp(\bm{u}_{i}^T \bm{u}_{s})} \numberthis\\
\bm{s} &= \sum_{i} \alpha_{i} \bm{h}_{i}
\end{align*}

\noindent\textbf{Output Layer.}
Each sentence of the thread is represented using a concatenation of the corresponding sentence and thread vectors.
Thus, both sentence- and thread-level context are taken into consideration when predicting if the sentence is in the summary.
We use a dense layer and a cross-entropy loss for the output.

Two additional improvements are crucial for the HAN models: 
1) \textbf{\textit{pretrain}}. The models are initially designed for text classification.
Using the thread vectors and thread category labels~\cite{Bhatia:2016}, we are able to pretrain the HAN models on a text classification task.
We hypothesize that the pretrained sentence and thread encoders are well-suited for the summarization task.
2) \textbf{\textit{redundancy removal}}. Supervised summarization models do not handle redundancy well.
Following~\cite{Cao:2017}, we apply a redundancy removal step, where sentences of high relevance are iteratively added to the summary and a sentence is added if it contains at least 50\% new bigrams that are not previously contained in the summary.

\section{Data}
\label{sec:data}

Having described the HAN models for summarization in the previous section, we next present our data.
We use forum threads collected by Bhatia et al.~\shortcite{Bhatia:2014} from \url{tripadvisor.com} and \url{ubuntuforums.org}.
The data contain respectively 83,075 and 113,277 threads from TripAdvisor and UbuntuForums. 
Among them, 1,480 and 1,174 threads have category labels~\cite{Bhatia:2016} and are used for model pretraining.
Bhatia et al.~\shortcite{Bhatia:2014} annotated 100 TripAdvisor threads with human summaries.
In this work we extend the summary annotation with 600 more threads, making a total of 700 threads.\footnote{The data is available at \url{http://tinyurl.com/jcqgcu8}}
We recruited six annotators and instructed them to read each thread and produce a summary of 10\% to 25\% of the original thread length.
They can use sentences in the thread or their own words. 
Two human summaries are created per thread.
We set aside 100 threads as a dev set and report results on the rest 600 threads.
In total, there are 34,033 sentences in the 600 threads.
A thread contains 10.5 posts and 56.2 sentences averagely. 


Further, we need to obtain sentence-level summary labels, where 1 means the sentence is in the gold-standard summary and 0 otherwise.
This is accomplished using an iterative greedy selection process.
Starting from an empty set, we add one sentence to the summary in each iteration such that the sentence produces the most improvement on ROUGE-1 scores~\cite{Lin:2004}.
The process stops if none could improve the ROUGE-1 scores, or if the summary has reached a pre-specified length limit of 20\% of the total words in the thread.
Note that, since there are two human summaries for every forum thread, ROUGE-1 scores measure the unigram overlap between the selected sentences and both of the human summaries. 
ROUGE 2.0 Java package was used for evaluation.

\begin{table*}
\setlength{\tabcolsep}{6pt}
\renewcommand{\arraystretch}{1.15}
\centering
\begin{small}
\begin{tabular}{l||r r r||r r r||r r r}
& \multicolumn{3}{c||}{\textbf{ROUGE-1}} & \multicolumn{3}{c||}{\textbf{ROUGE-2}} & \multicolumn{3}{c}{\textbf{Sentence-Level}}\\
\textbf{System} & \multicolumn{1}{c}{\textbf{R} (\%)} & \multicolumn{1}{c}{\textbf{P} (\%)} & \multicolumn{1}{c||}{\textbf{F} (\%)} & \multicolumn{1}{c}{\textbf{R} (\%)} & \multicolumn{1}{c}{\textbf{P} (\%)} & \multicolumn{1}{c||}{\textbf{F} (\%)} & \multicolumn{1}{c}{\textbf{R} (\%)} & \multicolumn{1}{c}{\textbf{P} (\%)} & \multicolumn{1}{c}{\textbf{F} (\%)}\\
\hline
\hline
\textbf{ILP} & 24.5 & 41.1 & 29.3$\pm$0.5 & 7.9 & 15.0 & 9.9$\pm$0.5 & 13.6 & 22.6 & 15.6$\pm$0.4\\
\textbf{Sum-Basic} & 28.4 & 44.4 & 33.1$\pm$0.5 & 8.5 & 15.6 & 10.4$\pm$0.4 & 14.7 & 22.9 & 16.7$\pm$0.5\\
\textbf{KL-Sum} & 39.5 & 34.6 & 35.5$\pm$0.5 & 13.0 & 12.7 & 12.3$\pm$0.5 & 15.2 & 21.1 & 16.3$\pm$0.5\\
\textbf{LexRank} & 42.1 & 39.5 & 38.7$\pm$0.5 & 14.7 & 15.3 & 14.2$\pm$0.5 & 14.3 & 21.5 & 16.0$\pm$0.5\\
\textbf{MEAD} & 45.5 & 36.5 & 38.5$\pm$ 0.5 & 17.9 & 14.9 & 15.4$\pm$0.5 & 27.8 & 29.2 & 26.8$\pm$0.5\\
\hline
\textbf{SVM} & 19.0 & 48.8 & 24.7$\pm$0.8 & 7.5 & 21.1 & 10.0$\pm$0.5 & 32.7 & 34.3 & 31.4$\pm$0.4\\
\textbf{LogReg} & 26.9 & 34.5 & 28.7$\pm$0.6 & 6.4 & 9.9 & 7.3$\pm$0.4 & 12.2 & 14.9 & 12.7$\pm$0.5\\
\textbf{LogReg}$^{\bm{r}}$ & 28.0 & 34.8 & 29.4$\pm$0.6 & 6.9 & 10.4 & 7.8$\pm$0.4 & 12.1 & 14.5 & 12.5$\pm$0.5\\
\hline
\hline
\textbf{HAN} & {\cellcolor[gray]{.8}} 31.0 & 42.8 & 33.7$\pm$0.7 & {\cellcolor[gray]{.8}} 11.2 & 17.8 & 12.7$\pm$0.5 & {\cellcolor[gray]{.8}} 26.9 & 34.1 & 32.4$\pm$0.5\\
\textbf{HAN+pretrainT} & {\cellcolor[gray]{.8}} 32.2 & 42.4 & 34.4$\pm$0.7 & {\cellcolor[gray]{.8}} 11.5 & 17.5 & 12.9$\pm$0.5 & {\cellcolor[gray]{.8}} 29.6 & 35.8 & 32.2$\pm$0.5\\
\textbf{HAN+pretrainU} & {\cellcolor[gray]{.8}} 32.1 & 42.1 & 33.8$\pm$0.7 & {\cellcolor[gray]{.8}} 11.6 & 17.6 & 12.9$\pm$0.5 & {\cellcolor[gray]{.8}} 30.1 & 35.6 & 32.3$\pm$0.5\\
\hline
\textbf{HAN}$^{\bm{r}}$ & {\cellcolor[gray]{.8}} 38.1 & 40.5 & \textbf{37.8}$\pm$0.5 & {\cellcolor[gray]{.8}} 14.0 & 17.1 & \textbf{14.7}$\pm$0.5 & {\cellcolor[gray]{.8}} 32.5 & 34.4 & \textbf{33.4}$\pm$0.5\\
\textbf{HAN+pretrainT}$^{\bm{r}}$ & {\cellcolor[gray]{.8}} 37.9 & 40.4 & \textbf{37.6}$\pm$0.5 & {\cellcolor[gray]{.8}} 13.5 & 16.8 & \textbf{14.4}$\pm$0.5 & {\cellcolor[gray]{.8}} 32.5 & 34.4 & \textbf{33.4}$\pm$0.5\\
\textbf{HAN+pretrainU}$^{\bm{r}}$ & {\cellcolor[gray]{.8}} 37.9 & 40.4 & \textbf{37.6}$\pm$0.5 & {\cellcolor[gray]{.8}} 13.6 & 16.9 & \textbf{14.4}$\pm$0.5 & {\cellcolor[gray]{.8}} 33.9 & 33.8 & \textbf{33.8}$\pm$0.5\\
\end{tabular}
\end{small}
\caption{Results of thread summarization. 
`HAN' models are our proposed approaches adapted from the hierarchical attention networks~\cite{Yang:2016}. 
The models can be pretrained using unlabeled threads from TripAdvisor (`T') and Ubuntuforum (`U'). 
$\bm{r}$ indicates a redundancy removal step is applied.
We report the variance of F-scores across all threads (`$\pm$').
A redundancy removal step improves recall scores (shown in gray) of the HAN models and boosts performance. 
}
\label{tab:results}
\vspace{-0.1in}
\end{table*}

\section{Experimental Setup}
\label{sec:experiments}

\noindent\textbf{Unsupervised baselines}.
Our proposed approach is compared against a range of unsupervised baselines, including 
1) \textsc{ILP}~\cite{Berg-Kirkpatrick:2011}, a baseline integer linear programming (ILP) framework implemented by~\cite{Boudin:2015};
2) \textsc{SumBasic}~\cite{Vanderwende:2007}, an approach that assumes words occurring frequently in a document cluster have a higher chance of being included in the summary;
3) \textsc{KL-Sum}, a method that adds sentences to the summary so long as it decreases the KL Divergence;
4) \textsc{LexRank}~\cite{Erkan:2004}, a graph-based summarization approach based on eigenvector centrality;
5) \textsc{Mead}~\cite{Radev:2004}, a centroid-based summarization system that scores sentences based on length, centroid, and position.

\vspace{0.05in}
\noindent\textbf{Supervised baselines}.
We implemented two supervised baselines that use SVM and logistic regression to predict if a sentence is in the summary.
We use the LIBLINEAR implementation~\cite{Fan:2008} where features include 1) cosine similarity of current sentence to the thread centroid, 2) relative sentence position within thread, 3) number of words in the sentence excluding stopwords, 4) max/avg/total TF-IDF scores of the consisting words.
The features are designed such that they carry similar information as achievable by the HAN models.
We use the 100-thread dev set for tuning hyperparameters. 
The optimal ones are `-c 0.1 -w1 5' for LogReg and `-c 10 -w1 5' for SVM.  

\vspace{0.05in}
\noindent\textbf{HAN configurations}.
The HAN models use RMSProp~\cite{Tieleman:2012} for parameter optimization, which has been shown to converge fast in sequence learning tasks.
The number of sentences per thread is set to 144 and number of words per sentence is 40.
We produce 200-dimension sentence vectors and 100-dimension thread vectors.
Dropout for word embeddings was 20\% and the output layer 50\%.

\vspace{0.05in}
\noindent\textbf{Evaluation metrics}.
ROUGE~\cite{Lin:2004} measures the n-gram overlap between system and human summaries. 
In this work we report ROUGE-1 and ROUGE-2 scores since these are metrics commonly used in the DUC and TAC competitions~\cite{Dang:2008}.
Additionally, we calculate the sentence-level precision, recall, and f-scores by comparing system prediction with gold-standard sentence labels.
All system summaries use a length threshold of 20\% thread words.

\section{Results}
The experimental results of all models are shown in Table \ref{tab:results}.
The HAN models are compared with a set of unsupervised (ILP, Sum-Basic, KL-Sum, LexRank, and MEAD) and supervised (SVM, LogReg) approaches.
We describe the observations below.

\begin{itemize}

\item First, HAN models appear to be more appealing than SVM and LogReg because there is less variation in program implementation, hence less effort is required to reproduce the results.
HAN models outperform both LogReg and SVM using the current set of features.
They yield higher precision scores than traditional models.

\item With respect to ROUGE scores, the HAN models outperform all supervised and unsupervised baselines except MEAD.
MEAD has been shown to perform well in previous studies~\cite{Luo:2016:NAACL} and it appears to handle redundancy removal exceptionally well.
The HAN models outperform MEAD in terms of sentence prediction.

\item Pretraining the HAN models, although intuitively promising, yields only comparable results with those without.
We suspect that there are not enough data to pretrain the models and that the thread classification task used to pretrain the HAN models may not be sophisticated enough to learn effective thread vectors.

\item We observe that the redundancy removal step is crucial for the HAN models to achieve outstanding results.
It helps improve the recall scores of both ROUGE and sentence prediction.
When redundancy removal was applied to LogReg, it produces only marginal improvement. 
This suggests that future work may need to consider principled ways of redundancy removal.
\end{itemize}

\vspace{-0.04in}
\section{Related Work}
\label{sec:related}

There has been some related work on email thread summarization~\cite{Rambow:2004,Wan:2004,Carenini:2008:ACL,Murray:2008,Oya:2014}.
Many of these are driven by the publicly available Enron email corpus~\cite{Klimt:2004} and other mailing lists.
Supervised approaches to email summarization draw on features such as sentence length, position, subject, sender/receiver, etc. Maximum entropy, SVM, CRF and variants~\cite{Ding:2008} are used as classifiers.
Further, Uthus and Aha~\shortcite{Uthus:2011} described the opportunities and challenges of summarizing military chats.
Giannakopoulos et al.~\shortcite{Giannakopoulos:2015} presented a shared task on summarizing the comments found on news providers.
We expect the human summaries created in this work will enable development of new approaches for thread summarization.

A recent strand of research is to model abstractive summarization (e.g., headline generation) as a sequence to sequence learning task~\cite{Rush:2015,Wiseman:2016,Nallapati:2016}.
The models use an encoder to read a large chunk of input text and a decoder to generate a sentence one word at a time.
Training the models require a large data collection where headlines are paired up with the first sentence of the articles.
In contrast, our approach focuses on developing effective sentence and thread encoders and require less training data. 

\section{Conclusion}
\label{sec:conclusion}

Supervised summarization approaches provide a promising avenue for scoring sentences.
We have developed a class of supervised models by adapting the hierarchical attention networks to forum thread summarization. We compare the model with a range of unsupervised and supervised summarization baselines.
Our experimental results demonstrate that the model performs better than most baselines and has the ability to capture contextual information with the recurrent structure.
In particular, we believe that the incorporation of a redundancy removal step to supervised models is the key contributor to the results.

\bibliography{forum,summ,fei}

\begin{thebibliography}{}

\bibitem[\protect\citeauthoryear{Anderson \bgroup et al\mbox.\egroup
  }{2012}]{Anderson:2012}
Anderson, A.; Huttenlocher, D.; Kleinberg, J.; and Leskovec, J.
\newblock 2012.
\newblock Discovering value from community activity on focused question
  answering sites: {A} case study of stack overflow.
\newblock In {\em Proceedings of the ACM SIGKDD Conference on Knowledge
  Discovery and Data Mining (KDD)}.

\bibitem[\protect\citeauthoryear{Bahdanau, Cho, and
  Bengio}{2015}]{Bahdanau:2015}
Bahdanau, D.; Cho, K.; and Bengio, Y.
\newblock 2015.
\newblock Neural machine translation by jointly learning to align and
  translate.
\newblock In {\em Proceedings of the International Conference on Learning
  Representations (ICLR)}.

\bibitem[\protect\citeauthoryear{Berg-Kirkpatrick, Gillick, and
  Klein}{2011}]{Berg-Kirkpatrick:2011}
Berg-Kirkpatrick, T.; Gillick, D.; and Klein, D.
\newblock 2011.
\newblock Jointly learning to extract and compress.
\newblock In {\em Proceedings of ACL}.

\bibitem[\protect\citeauthoryear{Bhatia, Biyani, and Mitra}{2014}]{Bhatia:2014}
Bhatia, S.; Biyani, P.; and Mitra, P.
\newblock 2014.
\newblock Summarizing online forum discussions -- {C}an dialog acts of
  individual messages help?
\newblock In {\em Proceedings of the Conference on Empirical Methods in Natural
  Language Processing (EMNLP)}.

\bibitem[\protect\citeauthoryear{Bhatia, Biyani, and Mitra}{2016}]{Bhatia:2016}
Bhatia, S.; Biyani, P.; and Mitra, P.
\newblock 2016.
\newblock Identifying the role of individual user messages in an online
  discussion and its applications in thread retrieval.
\newblock {\em Journal of the Association for Information Science and
  Technology (JASIST)} 67(2):276--288.

\bibitem[\protect\citeauthoryear{Boudin, Mougard, and
  Favre}{2015}]{Boudin:2015}
Boudin, F.; Mougard, H.; and Favre, B.
\newblock 2015.
\newblock Concept-based summarization using integer linear programming: {F}rom
  concept pruning to multiple optimal solutions.
\newblock In {\em Proceedings of the Conference on Empirical Methods in Natural
  Language Processing (EMNLP)}.

\bibitem[\protect\citeauthoryear{Cao \bgroup et al\mbox.\egroup
  }{2017}]{Cao:2017}
Cao, Z.; Li, W.; Li, S.; and Wei, F.
\newblock 2017.
\newblock Improving multi-document summarization via text classification.
\newblock In {\em Proceedings of the Association for the Advancement of
  Artificial Intelligence (AAAI)}.

\bibitem[\protect\citeauthoryear{Carenini, Ng, and
  Zhou}{2008}]{Carenini:2008:ACL}
Carenini, G.; Ng, R.~T.; and Zhou, X.
\newblock 2008.
\newblock Summarizing emails with conversational cohesion and subjectivity.
\newblock In {\em Proceedings of the Annual Meeting of the Association for
  Computational Linguistics (ACL)}.

\bibitem[\protect\citeauthoryear{Chen, Bolton, and Manning}{2016}]{Chen:2016}
Chen, D.; Bolton, J.; and Manning, C.~D.
\newblock 2016.
\newblock A thorough examination of the cnn/daily mail reading comprehension
  task.
\newblock In {\em Proceedings of ACL}.

\bibitem[\protect\citeauthoryear{Chung \bgroup et al\mbox.\egroup
  }{2014}]{Chung:2014}
Chung, J.; Gulcehre, C.; Cho, K.; and Bengio, Y.
\newblock 2014.
\newblock Empirical evaluation of gated recurrent neural networks on sequence
  modeling.
\newblock In {\em Proceedings of NIPS 2014 Workshop on Deep Learning}.

\bibitem[\protect\citeauthoryear{Dang and Owczarzak}{2008}]{Dang:2008}
Dang, H.~T., and Owczarzak, K.
\newblock 2008.
\newblock Overview of the {TAC} 2008 update summarization task.
\newblock In {\em Proceedings of Text Analysis Conference (TAC)}.

\bibitem[\protect\citeauthoryear{Ding and Jiang}{2015}]{Ding:2015}
Ding, Y., and Jiang, J.
\newblock 2015.
\newblock Towards opinion summarization from online forums.
\newblock In {\em Proceedings of the International Conference Recent Advances
  in Natural Language Processing (RANLP)}.

\bibitem[\protect\citeauthoryear{Ding \bgroup et al\mbox.\egroup
  }{2008}]{Ding:2008}
Ding, S.; Cong, G.; Lin, C.-Y.; and Zhu, X.
\newblock 2008.
\newblock Using conditional random fields to extract contexts and answers of
  questions from online forums.
\newblock In {\em Proceedings of ACL}.

\bibitem[\protect\citeauthoryear{Erkan and Radev}{2004}]{Erkan:2004}
Erkan, G., and Radev, D.~R.
\newblock 2004.
\newblock {LexRank}: {G}raph-based lexical centrality as salience in text
  summarization.
\newblock {\em Journal of Artificial Intelligence Research}.

\bibitem[\protect\citeauthoryear{Fan \bgroup et al\mbox.\egroup
  }{2008}]{Fan:2008}
Fan, R.-E.; Chang, K.-W.; Hsieh, C.-J.; Wang, X.-R.; and Lin, C.-J.
\newblock 2008.
\newblock {LIBLINEAR}: {A} library for large linear classification.
\newblock {\em Journal of Machine Learning Research} 9:1871--1874.

\bibitem[\protect\citeauthoryear{Giannakopoulos \bgroup et al\mbox.\egroup
  }{2015}]{Giannakopoulos:2015}
Giannakopoulos, G.; Kubina, J.; Conroy, J.~M.; Steinberger, J.; Favre, B.;
  Kabadjov, M.; Kruschwitz, U.; and Poesio, M.
\newblock 2015.
\newblock {MultiLing} 2015: {M}ultilingual summarization of single and
  multi-documents, on-line fora, and call-center conversations.
\newblock In {\em Proceedings of SIGDIAL}.

\bibitem[\protect\citeauthoryear{Hochreiter and
  Schmidhuber}{1997}]{hochreiter-97}
Hochreiter, S., and Schmidhuber, J.
\newblock 1997.
\newblock Long short-term memory.
\newblock {\em Neural computation} 9(8):1735--1780.

\bibitem[\protect\citeauthoryear{Klimt and Yang}{2004}]{Klimt:2004}
Klimt, B., and Yang, Y.
\newblock 2004.
\newblock The enron corpus: {A} new dataset for email classification research.
\newblock In {\em Proceedings of ECML}.

\bibitem[\protect\citeauthoryear{Li, Luong, and Jurafsky}{2015}]{Li:2015}
Li, J.; Luong, M.-T.; and Jurafsky, D.
\newblock 2015.
\newblock A hierarchical neural autoencoder for paragraphs and documents.
\newblock In {\em Proceedings of the 53rd Annual Meeting of the Association for
  Computational Linguistics (ACL)}.

\bibitem[\protect\citeauthoryear{Lin}{2004}]{Lin:2004}
Lin, C.-Y.
\newblock 2004.
\newblock {ROUGE}: a package for automatic evaluation of summaries.
\newblock In {\em Proceedings of ACL Workshop on Text Summarization Branches
  Out}.

\bibitem[\protect\citeauthoryear{Luo \bgroup et al\mbox.\egroup
  }{2016}]{Luo:2016:NAACL}
Luo, W.; Liu, F.; Liu, Z.; and Litman, D.
\newblock 2016.
\newblock Automatic summarization of student course feedback.
\newblock In {\em Proceedings of NAACL}.

\bibitem[\protect\citeauthoryear{Mikolov \bgroup et al\mbox.\egroup
  }{2013}]{mikolov-13}
Mikolov, T.; Chen, K.; Corrado, G.; and Dean, J.
\newblock 2013.
\newblock Efficient estimation of word representations in vector space.
\newblock {\em arXiv preprint arXiv:1301.3781}.

\bibitem[\protect\citeauthoryear{Murray and Carenini}{2008}]{Murray:2008}
Murray, G., and Carenini, G.
\newblock 2008.
\newblock Summarizing spoken and written conversations.
\newblock In {\em Proceedings of the Conference on Empirical Methods in Natural
  Language Processing (EMNLP)}.

\bibitem[\protect\citeauthoryear{Nallapati \bgroup et al\mbox.\egroup
  }{2016}]{Nallapati:2016}
Nallapati, R.; Zhou, B.; dos Santos, C.; Gulcehre, C.; and Xiang, B.
\newblock 2016.
\newblock Abstractive text summarization using sequence-to-sequence {RNNs} and
  beyond.
\newblock In {\em Proceedings of the 20th SIGNLL Conference on Computational
  Natural Language Learning (CoNLL)}.

\bibitem[\protect\citeauthoryear{Oya and Carenini}{2014}]{Oya:2014}
Oya, T., and Carenini, G.
\newblock 2014.
\newblock Extractive summarization and dialogue act modeling on email threads:
  {A}n integrated probabilistic approach.
\newblock In {\em Proceedings of the Annual SIGdial Meeting on Discourse and
  Dialogue (SIGDIAL)}.

\bibitem[\protect\citeauthoryear{Radev \bgroup et al\mbox.\egroup
  }{2004}]{Radev:2004}
Radev, D.~R.; Jing, H.; Sty\'{s}, M.; and Tam, D.
\newblock 2004.
\newblock Centroid-based summarization of multiple documents.
\newblock {\em Information Processing and Management} 40(6):919--938.

\bibitem[\protect\citeauthoryear{Rambow \bgroup et al\mbox.\egroup
  }{2004}]{Rambow:2004}
Rambow, O.; Shrestha, L.; Chen, J.; and Lauridsen, C.
\newblock 2004.
\newblock Summarizing email threads.
\newblock In {\em Proceedings of the North American Chapter of the Association
  for Computational Linguistics (NAACL)}.

\bibitem[\protect\citeauthoryear{Ren \bgroup et al\mbox.\egroup
  }{2011}]{Ren:2011}
Ren, Z.; Ma, J.; Wang, S.; and Liu, Y.
\newblock 2011.
\newblock Summarizing web forum threads based on a latent topic propagation
  process.
\newblock In {\em Proceedings of the 20th ACM International Conference on
  Information and Knowledge Management (CIKM)}.

\bibitem[\protect\citeauthoryear{Rush, Chopra, and Weston}{2015}]{Rush:2015}
Rush, A.~M.; Chopra, S.; and Weston, J.
\newblock 2015.
\newblock A neural attention model for sentence summarization.
\newblock In {\em Proceedings of the Conference on Empirical Methods in Natural
  Language Processing (EMNLP)}.

\bibitem[\protect\citeauthoryear{Stephen and Galak}{2012}]{Stephen:2012}
Stephen, A.~T., and Galak, J.
\newblock 2012.
\newblock The effects of traditional and social earned media on sales: {A}
  study of a microlending marketplace.
\newblock {\em Journal of Marketing Research} 49.

\bibitem[\protect\citeauthoryear{Tieleman and Hinton}{2012}]{Tieleman:2012}
Tieleman, T., and Hinton, G.
\newblock 2012.
\newblock {Lecture 6.5---RmsProp: Divide the gradient by a running average of
  its recent magnitude}.
\newblock COURSERA: Neural Networks for Machine Learning.

\bibitem[\protect\citeauthoryear{Uthus and Aha}{2011}]{Uthus:2011}
Uthus, D.~C., and Aha, D.~W.
\newblock 2011.
\newblock Plans toward automated chat summarization.
\newblock In {\em Proceedings of the ACL Workshop on Automatic Summarization
  for Different Genres, Media, and Languages}.

\bibitem[\protect\citeauthoryear{Vanderwende \bgroup et al\mbox.\egroup
  }{2007}]{Vanderwende:2007}
Vanderwende, L.; Suzuki, H.; Brockett, C.; and Nenkova, A.
\newblock 2007.
\newblock Beyond {SumBasic}: {T}ask-focused summarization with sentence
  simplification and lexical expansion.
\newblock {\em Information Processing and Management} 43(6):1606--1618.

\bibitem[\protect\citeauthoryear{Wan and McKeown}{2004}]{Wan:2004}
Wan, S., and McKeown, K.
\newblock 2004.
\newblock Generating overview summaries of ongoing email thread discussions.
\newblock In {\em Proceedings of the 20th International Conference on
  Computational Linguistics (COLING)}.

\bibitem[\protect\citeauthoryear{Wiseman and Rush}{2016}]{Wiseman:2016}
Wiseman, S., and Rush, A.~M.
\newblock 2016.
\newblock Sequence-to-sequence learning as beam-search opimization.
\newblock In {\em Proceedings of Empirical Methods on Natural Language
  Processing (EMNLP)}.

\bibitem[\protect\citeauthoryear{Yang \bgroup et al\mbox.\egroup
  }{2016}]{Yang:2016}
Yang, Z.; Yang, D.; Dyer, C.; He, X.; Smola, A.; and Hovy, E.
\newblock 2016.
\newblock Hierarchical attention networks for document classification.
\newblock In {\em Proceedings of the North American Chapter of the Association
  for Computational Linguistics (NAACL)}.

\bibitem[\protect\citeauthoryear{Yin, Ebert, and Schutze}{2016}]{Yin:2016}
Yin, W.; Ebert, S.; and Schutze, H.
\newblock 2016.
\newblock Attention-based convolutional neural network for machine
  comprehension.
\newblock In {\em Proceedings of the NAACL Workshop on Human-Computer Question
  Answering}.

\end{thebibliography}
\bibliographystyle{aaai}

\end{document}